\pdfoutput=1
\documentclass[sigconf, screen]{acmart}
\usepackage{algorithmic}
\usepackage{algorithm}
\usepackage{etoolbox}
\usepackage{multirow}
\usepackage{multicol}
\usepackage{graphicx}
\apptocmd{\sloppy}{\hbadness 10000\relax}{}{}
\usepackage{mathtools}

\AtBeginDocument{%
  }

\setcopyright{acmlicensed}
\copyrightyear{2018}
\acmYear{2018}
\acmDOI{XXXXXXX.XXXXXXX}

\acmISBN{978-1-4503-XXXX-X/2018/06}

\renewcommand\footnotetextcopyrightpermission[1]{}
\begin{document}

\title{Multi-modal Collaborative Optimization and  Expansion Network for Event-assisted Single-eye Expression Recognition}

\author{Runduo Han, Xiuping Liu, Shangxuan Yi, Yi Zhang, Hongchen Tan}

\begin{abstract}
In this paper,  we  proposed  a Multi-modal Collaborative Optimization and Expansion Network (MCO-E Net),  to use event modalities to resist challenges such as low light, high exposure, and high dynamic range in single-eye expression recognition tasks.
The MCO-E Net introduces two innovative designs: Multi-modal Collaborative Optimization Mamba (MCO-Mamba) and Heterogeneous Collaborative and Expansion Mixture-of-Experts (HCE-MoE).
MCO-Mamba, building upon Mamba, leverages dual-modal information to jointly optimize the model, facilitating collaborative interaction and fusion of modal semantics. This approach encourages the model to balance the learning of both modalities and harness their respective strengths.
HCE-MoE, on the other hand, employs a dynamic routing mechanism to distribute structurally varied experts (deep, attention, and focal), fostering collaborative learning of complementary semantics. This heterogeneous architecture systematically integrates diverse feature extraction paradigms to comprehensively capture expression semantics.
Extensive experiments demonstrate that our proposed network achieves competitive performance in the task of single-eye expression recognition, especially under poor lighting conditions. 
Code is available at \url{https://github.com/hrdhrd/MCO-E-Net}
\end{abstract}




\maketitle

\section{INTRODUCTION}
\label{sec:intro}
Single-eye Expression Recognition is an emerging computer vision technique that utilizes visual sensors to capture and analyze single-eye movement patterns.  
This innovative approach principal applications encompassing driver health detection, close-range human-computer interaction and other tasks. 
Compared with conventional facial expression recognition, this ocular-centric approach demonstrates inherent advantages in privacy protection and partial occlusion robustness. 
So, some  methods~\cite{8658392, barros2021ieyesyouimpact, 10.1145/3386901.3388917} have achieved excellent performance by strengthening the ability to extract local features (such as eye movement, pupil contraction, and eyebrow changes).
Nevertheless, performance degradation persists under challenging illumination scenarios, including low-light conditions, high-dynamic-range environments, and overexposure scenarios.
To address challenging lighting conditions, existing approaches have incorporated infrared imaging~\cite{10.1145/3386901.3388917} or depth sensing technologies~\cite{9102419, siddiqi2014depth}. 
However, these modalities fundamentally fail to capture and emphasize critical texture details and micro-movements around the eyes.

Event cameras, as neuromorphic vision sensors, produce asynchronous outputs precisely recording temporal coordinates, spatial positions, and polarity states of light changes. 
Subsequently, event camera can better capture extreme illumination changes around the eyes and subtle dynamic textures through ultra-high temporal resolution. 
However, compared to the RGB modality, the event stream semantics is extremely sparse. It is difficult for the model to mine sufficient discriminative semantics from the event stream alone. 
So, combining the advantages of RGB  and Event modalities is more conducive to capturing high-quality expression discrimination semantics. 
However,  the distinct differences between Event streams and RGB sequences in terms of generation mechanisms, spatiotemporal data representations, and semantic richness create a modality gap between the two.  In addition, the two modalities belong to long-term data, which further increases the difficulty of collaborative modeling.

SEEN~\cite{Zhang_2023} established the first framework combining Event and RGB modalities. 
MSKD~\cite{ijcai2024p350} introduced a cross-modal distillation framework that enables knowledge transfer from Event-RGB enhanced teacher networks to RGB-only student models. 
However, they ignore the fine-grained alignment and fusion of event modalities and RGB modalities.
Therefore, we try to explore the collaborative modeling methods of two types of long-sequence event data and RGB data.
Building upon the perception and capture of high-quality spatiotemporal semantics, Mamba-based Methods~\cite{10902569, dong2024fusionmambacrossmodalityobjectdetection, 10783777, liu2024cross} demonstrate superior performance in multimodal semantic collaborative perception and modeling.  
However, the direct concatenation of dual-modality semantics still leads to misalignment and inconsistencies in spatiotemporal semantics.  
Besides, some methods~\cite{wan2024sigmasiamesemambanetwork, wang2024hmoeheterogeneousmixtureexperts, 10889975} attempt to adopt an alternating optimization model parameter space to align the semantic distributions of the two modalities. 
This is because the ultimate objective of deep models is to fit or learn the distribution of data. Consequently, the outcome of efficient collaborative modeling between two modalities is that the two modal distributions learned by the model become aligned.
However, in these methods~\cite{wan2024sigmasiamesemambanetwork, wang2024hmoeheterogeneousmixtureexperts, 10889975}, the two modes of information remain independent and lack direct participation of modality data in the process of trying to learn the common distribution.

In addition,  there are common and unique semantics in the discrimination of different expressions.  
The semantic discrimination provided by the eye region alone is very limited. 
Based on the traditional single-branch deep model, the unique semantics of different expressions are easily coupled with each other, thereby reducing the discriminative ability. 
Mixture of Experts (MoE)~\cite{6797059, shazeer2017outrageouslylargeneuralnetworks} is a model that combines multiple sub-model experts and a gating mechanism to perceive and encode diverse semantic representations from different perspectives,  to alleviate the semantic coupling problem of traditional deep models.
Conventional MoE implementations face inherent limitations due to their architecturally homogeneous experts with equivalent representational capacities. 
This structural uniformity induces overlapping feature learning across experts, which undermines their potential for specialization. 
Recently, HMoE~\cite{wang2024hmoeheterogeneousmixtureexperts} reveals such homogeneity constrains models' ability to address heterogeneous complexity demands.

To solve the  above two  issues, we proposed  a Multi-modal Collaborative Optimization and Expansion Network (MCO-E Net). 
The MCO-E Net contains two novel designs:  Multi-modal Collaborative Optimization  Mamba (MCO-Mamba),  Heterogeneous Collaborative and Expansion MoE (HCE-MoE). 
In  MCO-Mamba,  based on Mamba, we use two-modal information to jointly optimize the model, and perform collaborative interaction and fusion of modal semantics to drive the model to balance the learning of two-modal semantics and capture the advantages of both modalities.
In the  HCE-MoE,  distributes structurally diversified experts (deep, attention and focal) through dynamic routing mechanism, enabling collaborative learning of complementary semantics. This heterogeneous architecture systematically combines diverse feature expertise knowledge extraction paradigms to capture comprehensive expression semantics. 
Contributions of this work are as follows: 
\begin{itemize}
\item[$\bullet$] We design the MCO-Mamba  to better   align and fuse  the Event and RGB modalities in a collaborative manner, aiming to capture high-quality expression descriptors.
\item[$\bullet$] We design the HCE-MoE  to enable collaborative learning of complementary visual representations. 
\item[$\bullet$] Extensive experiments demonstrate that our MCO-E Net achieves  comptitive performance in event-based single-eye expression recognition. 
\end{itemize}

\section{RELATED WORKS}
\label{sec:related}
\subsection{Expression Recognition}

Current facial expression recognition methods~\cite{zheng2023posterpyramidcrossfusiontransformer, lee2020multi, zhang2023weakly, li2023decoupled, 10820048} predominantly rely on RGB data but exhibit sensitivity to lighting variations and occlusions. Subsequently, MRAN~\cite{lee2020multi} processes synchronized color, depth, and thermal streams via spatiotemporal attention mechanisms, while DMD~\cite{li2023decoupled} employs graph-based distillation units to optimize cross-modal integration. 
However, these methods face practical challenges such as privacy concerns and hardware constraints.
To relieve these issues, Zhang et al.~\cite{Zhang_2023}  and   MSKD~\cite{ijcai2024p350} are eye expression recognition methods that use event streams to protect privacy and resist the challenges of poor lighting conditions. Inspired by them~\cite{Zhang_2023, ijcai2024p350}, we further design an efficient RGB and event modality collaborative modeling mechanism to mine and fuse the semantic advantages of the two modalities.

\subsection{Mamba}
The recently introduced Mamba~\cite{gu2023mamba} architecture, integrating State Space Models (SSMs)~\cite{gu2022efficientlymodelinglongsequences, gu2021combiningrecurrentconvolutionalcontinuoustime} from control theory~\cite{5311910}, combines fast inference with linear sequence-length scaling, enabling efficient long-range dependency modeling. Its vision-specific variants~\cite{zhu2024visionmambaefficientvisual, liu2024vmambavisualstatespace, li2024videomambastatespacemodel} and multimodal extensions~\cite{10902569, dong2024fusionmambacrossmodalityobjectdetection, 10783777, liu2024cross} demonstrate strengths in processing heterogeneous data (video, audio, language). However, existing methods rely on direct feature concatenation without addressing modality gaps. Recent works like Sigma~\cite{wan2024sigmasiamesemambanetwork} (partial SSM parameter exchange), MSFMamba~\cite{gao2025msfmambamultiscalefeaturefusion} (full parameter exchange), and DepMamba~\cite{10889975} (selective parameter sharing) attempt cross-modal alignment but enforce static fusion strategies, risking modality-specific feature degradation or shared-specific imbalance. To resolve this, we propose an adaptive coupling mechanism that dynamically balances modality-shared and modality-specific features during interaction, enabling task-driven cross-modal fusion.

\subsection{Mixture-of-Experts}
The Mixture of Experts (MoE) framework, developed by Jacobs et al.~\cite{6797059} , enables specialized components to autonomously process segmented data domains and then integrate them uniformly. 
On this basis, SMoE~\cite{shazeer2017outrageouslylargeneuralnetworks} proposed Sparsely-Gated Mixture-of-Experts, which employs a gating network for expert selection and proposes a Top-$K$ routing strategy, that is, selecting the $K$ experts with the highest probability. 
Zhou et al.~\cite{zhou2022mixture} proposed expert choice routing, change the routing method from selecting top-k experts for each token to selecting top-k tokens for each expert. 
Hard MoE~\cite{gross2017hardmixturesexpertslarge}, employing a single decoding layer, demonstrates efficient trainability while achieving competitive performance on large-scale hashtag prediction benchmarks. 
HMoE~\cite{wang2024hmoeheterogeneousmixtureexperts} solves the expert homogeneity problem by changing the parameter dimension size of the expert sub-network, but this modulation mechanism fundamentally retains the
unified structure of the experts. 
Different from these prior works, we exploit structural heterogeneity to design a Mixture-of-Experts that can extract different expertise through experts with different structures.

\section{METHODOLOGY}
\definecolor{shapecolor}{rgb}{0.0,0.5,0.0}
\begin{figure*}[ht]
    \centering
    \includegraphics[width=0.96\linewidth]{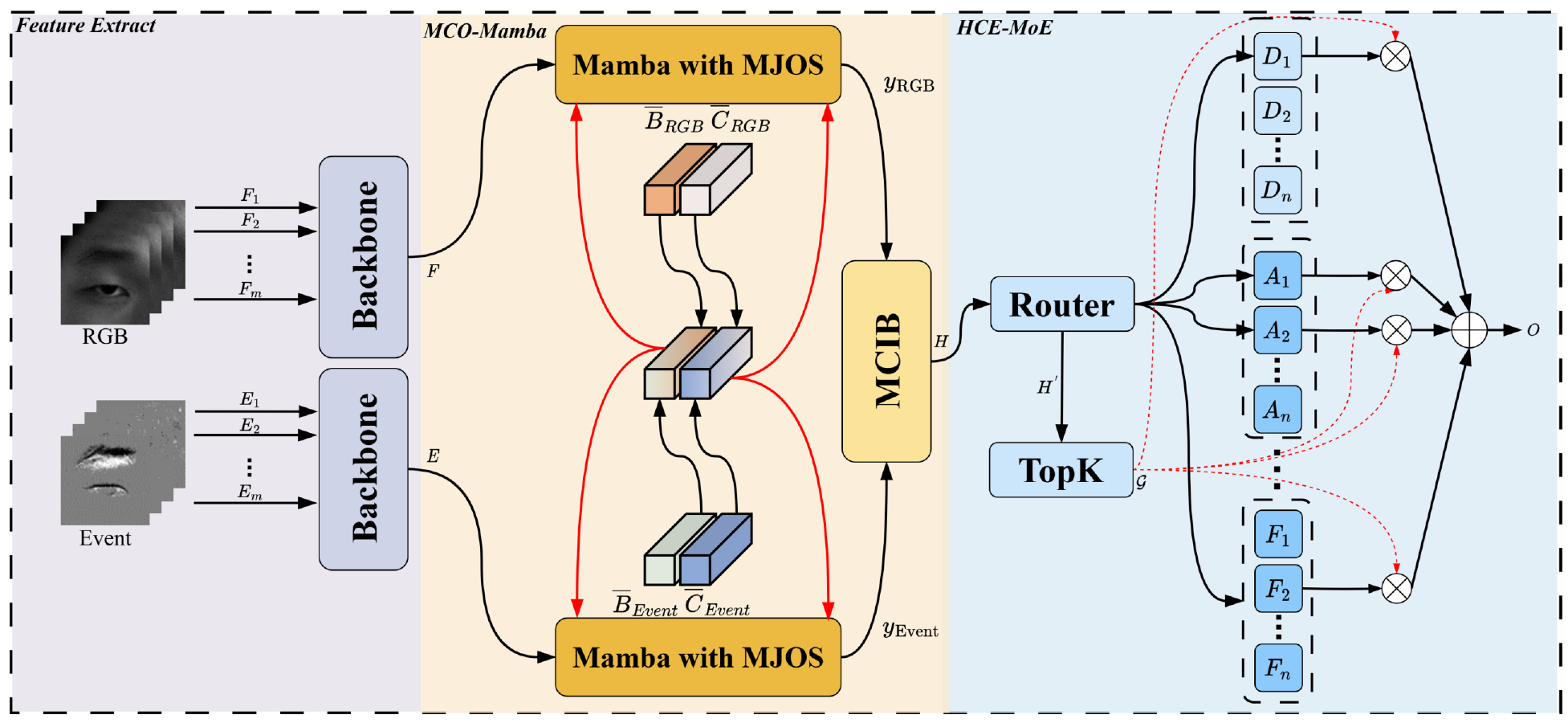}
    \caption{Overall architecture of our proposed MCO-E Net. The Network contains two novel designs: Multi-modal Collaborative Optimization Mamba (MCO-Mamba) and Heterogeneous Collaborative and Expansion Mixture of Experts (HCE-MoE).   }
    \label{fig:pipeline}
\end{figure*}

\subsection{Basic Information about Event Data}
Following~\cite{ahmad2023personreidentificationidentificationevent}, we  also convert the asynchronous event data into a voxel grid. 
The event sequence is represented as $E_1, E_2, \cdots, E_m$, where the dimension of each event tensor $E_i \in \mathbb{R}^{H\times W\times B}$. 
In the event tensor $E_i$,  the spatio-temporal coordinates, $x_k \in H$, $y_k \in W$, $t_b \in (B-1)$, lie on a voxel grid such that $x_k \in \{ 1, 2, ... , H \}, y_k \in \{ 1, 2, ..., W \}$, and $t_b \in \{ t_0, t_0 +\bigtriangleup t, ..., t_0 + (B-1) \bigtriangleup t \}$, where $t_0$ is the first time stamp, $\bigtriangleup t$ is the bin size, and $B-1$ is the number of temporal bins and $W, H$ are the sensor width and height.

\subsection{Overview}
Overall architecture of our MCO-E Net is shown in Fig.~\ref{fig:pipeline}. 
Input information  contains: RGB sequences $F_{i} \in \mathbb{R}^{H\times W \times C},i=1,2,\cdots,m$ and Event sequences $E_{i} \in \mathbb{R}^{H\times W \times B},i=1,2,\cdots,m$, where $H,W,C$ and $B$ represent height, width, RGB channels and Event channels. 
\textbf{Firstly}, the RGB sequences and  Event sequences are fed into their respective backbones; Each tensor is processed by a ResNet-18~\cite{he2015deepresiduallearningimage} for feature extraction, and then all features are concatenated together to obtain $F \in \mathbb{R}^{M\times E}$ and $E\in \mathbb{R}^{M\times E}$, $E$ is the feature dimension, $M$ is sequence length. 
\textbf{Next},  features $F \in \mathbb{R}^{M \times E}$ and $E\in \mathbb{R}^{M \times E}$ undergo our Multi-modal Collaborative Optimization  Mamba (MCO-Mamba), jointly optimizing the model and performing collaborative interaction and fusion of modal semantics. 
\textbf{Finally}, the fused representation from  MCO-Mamba is processed through our Heterogeneous Collaborative and Expansion MoE (HCE-MoE). 
The HCE-MoE  combines diverse feature expertise knowledge extraction paradigms to capture comprehensive expression semantics.

\subsection{MCO-Mamba}
\begin{figure}[h]
    \centering
    \includegraphics[width=\linewidth]{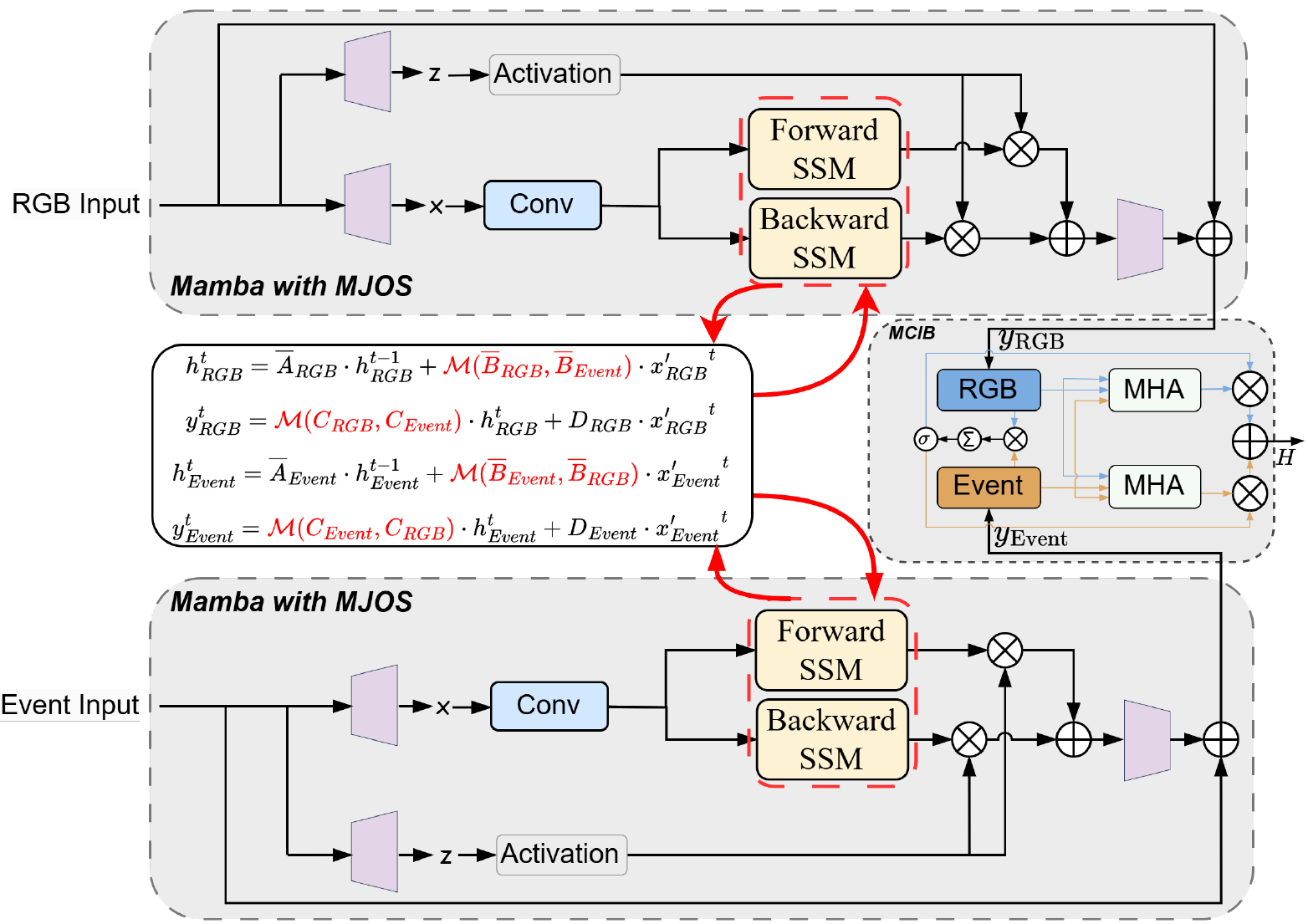}
    \caption{Architectures of  proposed MCO-Mamba. We first jointly optimize the model using Event and RGB modalities to drive the model to balance the learning of the two modal distributions; Next, we model the collaborative interaction between the two modalities to leverage their respective strengths and obtain high-quality expression descriptors. }
    \label{fig:mamba}
\end{figure}

As mentioned in Intro.~\ref{sec:intro}, the differences in information generation mechanism, representation form, and semantic richness between event streams and RGB data sequences lead to a modal semantic gap between the two modalities; In addition,  two modes have the characteristics of long sequences, which undoubtedly adds adjustment to the collaborative modeling of modes. 
To relieve two  sub-issues, we leverages the advantages of Mamba for long-sequence data modeling to design a multi-modal joint optimization and interactive representation model,  to fully leverage the advantages of both modalities and capture high-quality expression descriptors.
To  this end, we   proposed  the  Multi-modal Collaborative Optimization  Mamba (MCO-Mamba).

Our proposed MCO-Mamba is shown in Fig.~\ref{fig:mamba}, which consists of two core components: 
(i)Multi-modal Joint Optimization Scheme (MJOS) for Mamba: we jointly optimize Mamba with two modalities of information to improve its modeling of long-sequence information and perception of cross-modal information, thereby building a semantic bridge between the two modalities.
(ii)Multi-modal Collaborative Interaction Block (MCIB): we further interactively represent and fuse the two modalities to achieve complementary advantages of the two modalities.

\subsubsection{MJOS for Mamba}
\label{subsub:PIMamba}
To efficiently perceive and capture high-quality semantics from long-sequence data, we use Mamba to encode the semantics of Event streams and RGB sequences. 
However, Mamba is not good at collaborative modeling of two-modal semantics.
Therefore, some methods~\cite{10902569, dong2024fusionmambacrossmodalityobjectdetection, 10783777, liu2024cross} integrate the semantics of Mamba after modeling, but this does not perceive the role of sequence information in the fusion of modal semantics. 
Other method~\cite{wan2024sigmasiamesemambanetwork, gao2025msfmambamultiscalefeaturefusion} are to alternately update the time series status information of Mamba. 
Although these methods can effectively perceive the temporal semantics of two modalities, they lack the joint participation of two-modal information. This is not conducive to the balance of two-mode semantic distribution and the effective capture of semantic meaning in deep models.
So, we use two-modal sequence data to jointly optimize the state information of Mamba.  Here, we use SSM with bidirectional scanning~\cite{zhu2024visionmambaefficientvisual} to capture the forward and backward dependencies respectively to ensure that the output of each position can simultaneously refer to the context of the entire sequence. 
We define it as Multi-modal Joint Optimization Scheme (MJOS) for Mamba.  Specifically, the design details of our MJOS  for Mamba are shown in Algorithm~\ref{alg:mamba}.

\begin{algorithm}[]
\caption{Process of Mamba with Multi-modal Joint Optimization Scheme (MJOS) }
\label{alg:mamba}
\footnotesize 
\renewcommand{\algorithmicrequire}{\textbf{Input:}}
\renewcommand{\algorithmicensure}{\textbf{Output:}}
\begin{algorithmic}[1]
  \STATE \textcolor{gray}{\scriptsize/* $\textcolor{shapecolor}{\mathtt{D}}$ is the hidden state dimension, $\textcolor{shapecolor}{\mathtt{E}}$ is expanded state dimension, $\textcolor{shapecolor}{\mathtt{N}}$ is SSM dimension, $\textcolor{shapecolor}{\mathtt{B}}$ is batch size and $\textcolor{shapecolor}{\mathtt{M}}$ is sequence length. */}
  \REQUIRE sequence $\mathbf{E}, \mathbf{F}{:}\textcolor{shapecolor}{(B,M,E)}$
  \ENSURE sequence $\mathbf{y_{RGB}}, \mathbf{y_{Event}}{:}\textcolor{shapecolor}{(B,M,E)}$
  
  \STATE $\mathbf{X_{RGB}},\mathbf{Z_{RGB}}{:}\textcolor{shapecolor}{(B,M,E)}\gets\mathbf{Lin^{X_{RGB}}}(\mathbf{F}),\mathbf{Lin^{Z_{RGB}}}(\mathbf{F})$ 
  \STATE $\mathbf{X_{Event}},\mathbf{Z_{Event}}{:}\textcolor{shapecolor}{(B,M,E)}\gets\mathbf{Lin^{X_{Event}}}(\mathbf{E}),\mathbf{Lin^{Z_{Event}}}(\mathbf{E})$
  \STATE $\mathbf{X'_{RGB}},\mathbf{X'_{Event}}{:}\textcolor{shapecolor}{(B,M,E)}\gets\mathbf{Conv1d}(\mathbf{X_{RGB}}),\mathbf{Conv1d}(\mathbf{X_{Event}})$
  
  \FOR{$\mathit{o}$ in \{forward,backward\}}
    \STATE \textcolor{gray}{\scriptsize/* Parameter initialization */}
    \STATE $\mathbf{A^o_{RGB},A^o_{Event}}{:}\textcolor{shapecolor}{(D,N)}\gets\text{Parameter initialization}$ 
    \STATE $\mathbf{B^o_{RGB},B^o_{Event}}{:}\textcolor{shapecolor}{(B,M,N)}\gets\mathbf{Lin^{B^o_{RGB}}}(\mathbf{X'_{RGB}}),\mathbf{Lin^{B^o_{Event}}}(\mathbf{X'_{Event}})$
    \STATE $\mathbf{B^o_{Fusion}}\gets\mathbf{Concat}(\mathbf{B^o_{RGB}},\mathbf{B^o_{Event}})$
    \STATE $\mathbf{B^{o'}_{RGB},B^{o'}_{Event}}{:}\textcolor{shapecolor}{(B,M,N)}\gets\mathbf{Lin^{B^{o'}_{RGB}}}(\mathbf{B^o_{Fusion}})+\mathbf{B^o_{RGB}},\mathbf{Lin^{B^{o'}_{Event}}}(\mathbf{B^o_{Fusion}})+\mathbf{B^o_{Event}}$
    \STATE $\mathbf{C^o_{RGB},C^o_{Event}}{:}\textcolor{shapecolor}{(B,M,N)}\gets\mathbf{Lin^{C^o_{RGB}}}(\mathbf{X'_{RGB}}),\mathbf{Lin^{C^o_{Event}}}(\mathbf{X'_{Event}})$
    \STATE $\mathbf{C^o_{Fusion}}\gets\mathbf{Concat}(\mathbf{C^o_{RGB}},\mathbf{C^o_{Event}})$
    \STATE $\mathbf{C^{o'}_{RGB},C^{o'}_{Event}}{:}\textcolor{shapecolor}{(B,M,N)}\gets\mathbf{Lin^{C^{o'}_{RGB}}}(\mathbf{C^o_{Fusion}})+\mathbf{C^o_{RGB}},\mathbf{Lin^{C^{o'}_{Event}}}(\mathbf{C^o_{Fusion}})+\mathbf{C^o_{Event}}$
    \STATE $\mathbf{D^o_{RGB},D^o_{Event}}{:}\textcolor{shapecolor}{(D)}\gets\mathbf{1}$
    
    \STATE \textcolor{gray}{\scriptsize/* Discretize */}
    \STATE $\mathbf{\Delta^o_{RGB}}{:}\textcolor{shapecolor}{(B,M,E)}\gets\log(1{+}\exp(\mathbf{Lin^{\Delta^o_{RGB}}}(\mathbf{X'_{RGB}}){+}\mathbf{Param^{\Delta^o_{RGB}}}))$ 
    \STATE $\mathbf{\Delta^o_{Event}}{:}\textcolor{shapecolor}{(B,M,E)}\gets\log(1{+}\exp(\mathbf{Lin^{\Delta^o_{Event}}}(\mathbf{X'_{Event}}){+}\mathbf{Param^{\Delta^o_{Event}}}))$
    \STATE $\overline{\mathbf{A}}^o_{RGB},\overline{\mathbf{B}}^{o'}_{RGB}\gets\mathbf{discretize}(\mathbf{\Delta}^o_{RGB},\mathbf{A^o_{RGB}},\mathbf{B^{o'}_{RGB}})$
    \STATE $\overline{\mathbf{A}}^o_{Event},\overline{\mathbf{B}}^{o'}_{Event}\gets\mathbf{discretize}(\mathbf{\Delta}^o_{Event},\mathbf{A^o_{Event}},\mathbf{B^{o'}_{Event}})$
    
    \STATE \textcolor{gray}{\scriptsize/* State Space Model */}
    \STATE $\mathbf{y^o_{RGB}}{:}\textcolor{shapecolor}{(B,M,E)}\gets\mathbf{SSM}(\overline{\mathbf{A}}^o_{RGB},\overline{\mathbf{B}}^{o'}_{RGB},\mathbf{C^{o'}_{RGB}},\mathbf{D^o_{RGB}})(\mathbf{x'_{RGB}})$
    \STATE $\mathbf{y^o_{Event}}{:}\textcolor{shapecolor}{(B,M,E)}\gets\mathbf{SSM}(\overline{\mathbf{A}}^o_{Event},\overline{\mathbf{B}}^{o'}_{Event},\mathbf{C^{o'}_{Event}},\mathbf{D^o_{Event}})(\mathbf{x'_{Event}})$
  \ENDFOR
  
  \STATE $\mathbf{y_{RGB}}\gets\mathbf{Lin^{y_{RGB}}}(\mathbf{Z_{RGB}}{\cdot}(\mathbf{y^{forward}_{RGB}}{+}\mathbf{y^{backward}_{RGB}}))$
  \STATE $\mathbf{y_{Event}}\gets\mathbf{Lin^{y_{Event}}}(\mathbf{Z_{Event}}{\cdot}(\mathbf{y^{forward}_{Event}}{+}\mathbf{y^{backward}_{Event}}))$
  \RETURN $\mathbf{y_{RGB}},\mathbf{y_{Event}}{:}\textcolor{shapecolor}{(B,M,E)}$
\end{algorithmic}
\end{algorithm}

As  shown  in Algorithm~\ref{alg:mamba}, at different times, we jointly optimize the state equation of Mamba using the RGB modality and Event modality (as  shown in Fig.~\ref{fig:mamba}), as detailed below: 
\begin{align}
    h_{\text{RGB}}^{t} &= \overline{A}_{\text{RGB}} \cdot h_{\text{RGB}}^{t-1} + \underbrace{\mathcal{M}(\overline{B}_{\text{RGB}}, \overline{B}_{\text{Event}})}_{\text{Joint Optimization}} \cdot {x^{'}_{\text{RGB}}}^{t} \label{eq:B-interaction} \\
    y_{\text{RGB}}^{t} &= \underbrace{\mathcal{M}(C_{\text{RGB}}, C_{\text{Event}}) }_{\text{Joint Optimization}} \cdot h_{\text{RGB}}^{t} + D_{\text{RGB}} \cdot {x^{'}_{\text{RGB}}}^{t} \label{eq:C-interaction} \\
    h_{\text{Event}}^{t} &= \overline{A}_{\text{Event}} \cdot h_{\text{Event}}^{t-1} +\underbrace{\mathcal{M}(\overline{B}_{\text{Event}}, \overline{B}_{\text{RGB}})}_{\text{Joint Optimization}} \cdot {x^{'}_{\text{Event}}}^{t} \\
    y_{\text{Event}}^{t} &= \underbrace{\mathcal{M}(C_{\text{Event}}, C_{\text{RGB}})  }_{\text{Joint Optimization}} \cdot h_{\text{Event}}^{t} + D_{\text{Event}} \cdot {x^{'}_{\text{Event}}}^{t} \label{eq:event-output}
\end{align}
The multi-modal joint optimization function $\mathcal{M}$ is defined as:
\begin{equation}
    \mathcal{M}(\mathcal{A},\mathcal{B}) = \big(W \cdot [\mathcal{A};\mathcal{B}] + b\big) \oplus \mathcal{A} 
\label{eq:M-function}
\end{equation}
where $[;]$ denotes concatenation and $\oplus$ represents element-wise addition. This formulation enables feature concatenation from both modalities followed by learnable projections ($W$, $b$), creating shared parameters that preserve modality-specific characteristics while establishing cross modality associations.

\subsubsection{MCIB}
In  addition,  we further  introduce the Multi-modal Collaborative Interaction Block (MCIB) to  interactively represent the two-modal semantics to achieve fine-grained matching of modal semantics and complementary advantages of semantic.
As shown in Fig.~\ref{fig:mamba}, the block processes input features $\mathbf{y}_{\text{RGB}}$ and $\mathbf{y}_{\text{Event}}$ through Multi-head attention mechanism, $\mathbf{y}_{\text{RGB}}$ and $\mathbf{y}_{\text{Event}}$ are each other's Query. 
This design allows RGB features to actively retrieve motion details in event streams with their own semantics as anchors. 
Event streams features reversely focus on the static scene semantics of RGB, forming closed-loop semantic enhancement. 
The specific calculation process is as follows: 
\begin{equation}
    \text{Attention}(Q,K,V) = \text{Softmax}\left(\frac{QK^{\top}}{\sqrt{d_k}}\right)V \label{eq:attention}
\end{equation}
\begin{equation}
    \text{head}_i = \text{Attention}(QW_i^Q,KW^K_i,VW_i^V)
\end{equation}
\begin{equation}
    \text{MHA}(Q,K,V) = \text{Concat}(\text{head}_1,\cdots , \text{head}_h)W^O
\end{equation}
where $d_k$ is the dimension of K, MHA is multi head attention. Following, we take the projection of two modalities as each other's Query: 

\begin{equation}
{\small
\begin{split}
    Q_{\text{RGB}} &= W_{Q}^{\text{RGB}} \mathbf{y}_{\text{Event}}, \quad 
    K_{\text{RGB}} = W_{K}^{\text{RGB}} \mathbf{y}_{\text{RGB}}, \quad 
    V_{\text{RGB}} = W_{V}^{\text{RGB}} \mathbf{y}_{\text{RGB}} \\
    Q_{\text{Event}} &= W_{Q}^{\text{Event}} \mathbf{y}_{\text{RGB}}, \quad 
    K_{\text{Event}} = W_{K}^{\text{Event}} \mathbf{y}_{\text{Event}}, \quad 
    V_{\text{Event}} = W_{K}^{\text{Event}} \mathbf{y}_{\text{Event}}
\end{split}
}
\end{equation}
\begin{equation}
    \mathbf{y}'_{\text{RGB}} = \text{MHA}(Q_{\text{RGB}},K_{\text{RGB}},V_{\text{RGB}}), 
\end{equation}
\begin{equation}
    \mathbf{y}'_{\text{Event}} = \text{MHA}(Q_{\text{Event}},K_{\text{Event}},V_{\text{Event}})
\end{equation}
where $\mathbf{y_{RGB}}$ and $\mathbf{y_{Event}}$ are input features of semantic fusion block, $W$ is learnable linear projection. 

To adaptively balance modality contributions, we design a gating mechanism:
\begin{equation}
    \alpha = \sigma(\sum{(\mathbf{y_{RGB}}\odot \mathbf{y_{Event}})})
\end{equation}

\begin{equation}
    H = \alpha \cdot \mathbf{y'_{RGB}} + (1-\alpha) \cdot \mathbf{y'_{Event}}
\end{equation}
where $\sigma$ denotes the sigmoid function and $\odot$ represents element-wise multiplication. 
This gating strategy maintains inter-modality equilibrium without introducing additional parameters and avoid biasing towards a certain mode, effectively reducing model complexity while preserving fusion flexibility.

\subsection{HCE-MoE}
\begin{figure}[h]
    \centering
    \includegraphics[width=\linewidth]{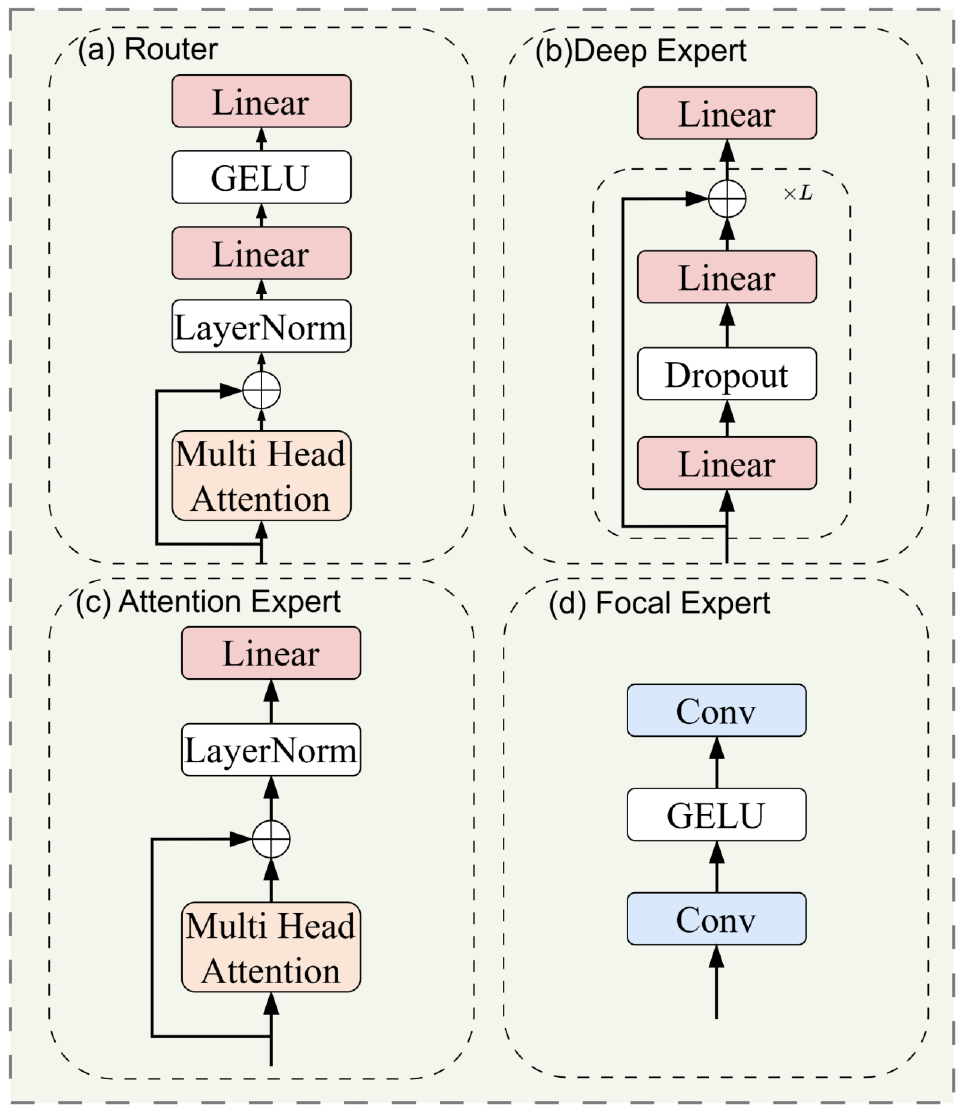}
    \caption{Detailed of the proposed Heterogeneous Collaborative and Expansion MoE (HCE-MoE). (a) Router with Attention. (b) Deep Expert. (c) Attention Expert. (d) Focal Expert. }
    \label{fig:moe}
\end{figure}

As described in Intro.~\ref{sec:intro},  we  try  to adopt Mixture-of-Experts (MoE)~\cite{6797059}  to dynamically selects  experts  to obtain diverse semantic representations.  
Compared to conventional MOE methods~\cite{wang2024hmoeheterogeneousmixtureexperts}, we design different types of expert models (i.e. Heterogeneous Experts) to try to reduce the overlap of semantic perception among multiple expert models and enhance semantic diversity.
Besides, most Mixture-of-Experts (MoE)~\cite{NEURIPS2022_3e67e84a, dai2024deepseekmoeultimateexpertspecialization} defines router as several linear projection layers, the static routing mechanism based on linear projection is difficult to deal with complex feature relationships, resulting in the inability to consider global context semantic when generating routing weights, which leads to rigid routing policy and leads to load imbalance. 
Inspired by~\cite{wu2024yuan, blecher2023moeatt} attention router, we further introduce attention-guided router  for MoE. 
So,  we  propose the Heterogeneous Collaborative and Expansion MoE (HCE-MoE). 
Proposed HCE-MoE  contains: \textbf{Router with Attention}, \textbf{Deep Expert}, \textbf{Attention Expert},  and \textbf{Focal Expert. }

\textbf{Router with Attention. }  Different from ~\cite{wu2024yuan, blecher2023moeatt} attention router, we  further combine the original features and attention-processed features and add nonlinear representations to improve router capabilities.
\begin{align}
    H' &= \mathcal{P}_2\left(\sigma\big(\mathcal{P}_1(\mathcal{LN}(\text{MHA}(H) \oplus H))\big)\right) \label{eq:router} \\
    \mathcal{G} &= \{(m_i, h_i)\}_{i=1}^k = \text{TopK}\left(\text{Softmax}(H')\right) \label{eq:gating}
\end{align}
where $H^{'} \in \mathbb{R}^{N_e}$, $N_e$ is number of experts. $\mathcal{P}_{1/2}$ are linear projections, $\text{MHA}(\cdot)$ denotes multi-head attention, and $\mathcal{G}$ contains selected expert indices $m_i$ with corresponding weights $h_i$, Top-K takes the first $k$ maximum value operation. 

Compared with traditional MoE routers, our structure can more carefully model the mapping relationship input to experts through hierarchical processing of $\mathcal{P}_{1} \to \sigma \to \mathcal{P}_{2}$, reducing routing confusion. 
Compared with existing methods of using attention routers~\cite{wu2024yuan, blecher2023moeatt}, we combine the output of MHA and the original input, the router can simultaneously utilize the original features without attention processing and the global context of attention output, avoiding information loss while fusing context information. In addition, we have added layer normalization and activation functions, improve the adaptability of the router to diversified input.

In addition, we introduce three specialized experts, namely: Deep Expert, Attention Expert and Focal Expert.

\begin{table*}[tb!]
	\caption{Comparison with SOTA methods on SSE dataset in terms of UAR, WAR, Normal, Overexposure, Low-Light and HDR. Best results are shown in bold. }
	\label{tab:SOTA}
	\resizebox{\textwidth}{!}
	{
		\begin{tabular}{cc|cc|cccc|ccccccc}
			\hline
			\multicolumn{2}{c|}{\multirow{2}{*}{Methods}}         & \multicolumn{2}{c|}{Metrics(\%)} & \multicolumn{4}{c|}{Accuracy under lighting conditions(\%)}   & \multicolumn{7}{c}{Accuracy of emotion classification(\%)}                                                    \\ \cline{3-15} 
			\multicolumn{2}{c|}{}                                 & WAR             & UAR            & Normal        & Overexposure  & Low-Light    & HDR           & Happy         & Sadness       & Anger         & Disgust       & Surprise      & Fear          & Neutral       \\ \hline
			\multicolumn{1}{c|}{Former DFER\cite{zhao2021former}}               & Face & 65.8            & 67.2           & 70.1          & 65.4          & 66.2          & 61.1          & 81.5          & 75.2          & 85.8          & 59.4          & 39.3          & 50.8          & 78.6          \\
			\multicolumn{1}{c|}{Former DFER w/o pre-train} & Face & 48.0            & 48.0           & 47.0          & 51.9          & 45.6          & 47.2          & 44.1          & 65.2          & 46.0          & 66.5          & 28.0          & 50.3          & 36.1          \\
			\multicolumn{1}{c|}{R(2+1)D\cite{tran2018closer}}                   & Face & 49.7            & 51.5           & 54.3          & 50.3          & 44.4          & 49.3          & 63.6          & 45.5          & 65.7          & 27.8          & 33.3          & 37.9          & 86.6          \\
			\multicolumn{1}{c|}{3D Resnet18\cite{hara2018can}}               & Face & 49.1            & 50.5           & 51.9          & 51.4          & 44.8          & 47.8          & 54.8          & 45.4          & 67.7          & 23.8          & 37.2          & 42.8          & 81.6          \\
			\multicolumn{1}{c|}{Resnet50 + GRU\cite{cho2014learning}}            & Face & 35.2            & 34.7           & 43.0          & 35.7          & 28.9          & 32.8          & 27.9          & 38.0          & 49.7          & 44.5          & 6.9           & 70.0          & 5.6           \\
			\multicolumn{1}{c|}{Resnet18 + LSTM\cite{6795963}}           & Face & 56.3            & 58.0           & 57.9          & 60.4          & 53.9          & 52.5          & 57.8          & 86.0          & 64.9          & 46.5          & 9.2           & 81.6          & 59.8          \\ \hline
			\multicolumn{1}{c|}{EMO\cite{wu2020emo}}                       & Eye  & 63.1            & 63.3           & 61.8          & 62.8          & 60.1          & 69.6          & 75.0          & 75.1          & 70.2          & 48.1          & 37.5          & 54.1          & 82.8          \\
			\multicolumn{1}{c|}{EMO w/o pre-train}         & Eye  & 53.2            & 53.3           & 46.1          & 60.2          & 55.5          & 58.9          & 62.0          & 73.2          & 60.1          & 38.7          & 25.7          & 48.0          & 65.3          \\
			\multicolumn{1}{c|}{Eyemotion\cite{Hickson2017EyemotionCF}}                 & Eye  & 78.8            & 79.5           & 79.0          & 81.8          & 81.5          & 72.5          & 74.3          & 85.5          & 79.5          & 74.3          & 69.1          & 79.2          & 94.5          \\
			\multicolumn{1}{c|}{Eyemotion w/o pre-train}   & Eye  & 75.9            & 77.2           & 77.8          & 75.9          & 79.8          & 69.7          & 79.6          & 85.7          & 81.2          & 71.2          & 54.7          & 71.6          & \textbf{96.4} \\
			\multicolumn{1}{c|}{SEEN~\cite{Zhang_2023}}                      & Eye  & 83.6            & 84.1           & 83.3          & 85.6          & 80.8          & 84.8          & 85.0          & 89.9          & 92.2 & 76.7          & 72.1 & 87.7          & 85.2          \\ 
                \multicolumn{1}{c|}{MSKD~\cite{ijcai2024p350}}         & Eye  & 86.2    &86.6          & 84.4 &89.1 &88.3 &82.7          &85.6 &91.7 &92.3 &79.0 &79.4 &88.0 &90.3          \\ 
			\multicolumn{1}{c|}{HI-Net~\cite{Han2024HierarchicalEI}}     & Eye  & 86.9   & 87.7  & 84.6 & 90.3 & 87.2 & 85.2 & 93.4 & 95.5 & 87.8          & 85.3 & 70.6          & \textbf{91.2} & 89.8          \\ \hline
                \multicolumn{1}{c|}{\textbf{Ours}}   & Eye  & \textbf{91.3}   & \textbf{91.9}  & \textbf{87.7} & \textbf{93.8} & \textbf{93.6} & \textbf{90.2} & \textbf{97.0} & \textbf{98.4} & \textbf{94.7} & \textbf{87.7} & \textbf{81.7} & 89.4 & 94.6          \\ \hline
	\end{tabular}}
\end{table*}

\textbf{Deep Expert. } Traditional expert typically employ $1-2$ layers, shallow networks strive to compose low-level features into high-level semantics, while narrow intermediate layers restrict the model's ability to handle complex feature. 
So, our deep expert implements progressive feature refinement through $L$ stacked transformation blocks, each of the $L$ layers expands the input dimension to $4\times$ its original size, creating a high-dimensional space for features. Subsequent compression back to the original dimension ensures compatibility with residual connections while preserving key information:
\begin{equation}
O_{D} = \mathcal{P}_3(\left( \operatorname*{\bigcirc}_{l=1}^L \left[ \mathcal{P}_2^{(l)} \circ \mathcal{D}^{(l)} \circ \mathcal{P}_1^{(l)} \right] \right)(H))
\end{equation}
where $H$ is input feature, $\mathcal{P}_1^{(l)}: \mathbb{R}^d \to \mathbb{R}^{4d}$ expands dimensions, $\mathcal{P}_2^{(l)}: \mathbb{R}^{4d} \to \mathbb{R}^{d}$ restore dimensions, $\mathcal{P}_3^{(l)}: \mathbb{R}^{d} \to \mathbb{R}^{\mathcal{J}}, \mathcal{J}$ is number of expression class, $\mathcal{D}$ denotes dropout, $l$ is $l-th$ layer, $L$ is depth and $ \operatorname*{\bigcirc}_{l=1}^L$ is nested operations from the first layer to the $L$ layer.

\textbf{Attention Expert. } Ordinary experts are usually composed of only fully connected layers, lack explicit sequence modeling capabilities, and are difficult to deal with scenarios that require global context. 
While deep-oriented architectures excel at hierarchical feature extraction, they lack input-adaptive feature weighting crucial for emotion-varying contexts. 
To address this, we design attention experts specializing in global contextual reasoning through multi-head self-attention. Each expert introduces MHA internally to enable it to capture dependencies in the input sequence, which is crucial for sequence processing tasks:
\begin{equation}
    O_{A} = \mathcal{P}(\mathcal{LN}(\text{MHA}(H) \oplus H))
\end{equation}
where $O_{A} \in \mathbb{R}^{\mathcal{J}}, \mathcal{J}$ is number of expression class, $\text{MHA}$ is Multi head attention, $\mathcal{LN}$ is Layer Normalization and $\mathcal{P}$ is linear projection. Add the output of MHA with the original input, retaining the underlying feature information to avoid information loss in networks. 
In addition, layer normalization is used after MHA and residual connection to improve overfitting and enhance model generalization. 

\textbf{Focal Expert. }
The efficacy of Single-eye expression recognition heavily relies on precise localization of micro-expression patterns (e.g., eye narrowing or brow furrowing). While global attention mechanisms excel at contextual modeling, they often overlook fine-grained spatial details critical for subtle emotion discrimination. To achieve this, we design Focal expert specializing in hierarchical local feature: 
\begin{equation}
    O_{C} = C_{1}(\sigma(C_{3}(H)))
\end{equation}
where $O_{C} \in \mathbb{R}^{\mathcal{J}}, \mathcal{J}$ is number of expression class, $C_{i}$ denotes convolution with kernel size $i$.

\textbf{Expert Integration. }
Final prediction integrate outputs from activated experts through weighted aggregation: 
\begin{equation}
    O = \sum_{(m_i, h_i) \in \mathcal{G}}h_i \cdot O_{m_i}(H)
\end{equation}
where $(m_i, h_i) \in \mathcal{G}$ calculated from router with attention, $\sum h_i = 1$ ensures normalized contributions, $O_{m_i}(\cdot)$ is the activated expert, maybe $O_D$, $O_A$, $O_C$.

\section{EXPERIMENT}
\subsection{Datasets and settings}

\textbf{Datasets. }Experiments utilize the SEE dataset~\cite{Zhang_2023}, the sole public benchmark for single-eye emotion recognition. 
It contains 2,405 event sequences (128,712 RGB frames with synchronized event streams). Following official protocols, data is split into training (1,638 sequences) and test (767 sequences)

\textbf{Evaluation Metrics. } Performance is measured via: 
UAR (Unweighted Average Recall): Arithmetic mean of per-class recalls, ensuring balanced assessment across classes. Critical for handling imbalanced data. 
WAR (Weighted Average Recall): Sample-size weighted recall average, it prevents majority classes from dominating the evaluation while ensuring adequate sensitivity to minority class recognition. 

\textbf{Implementation Details. } We trained our network for 200 epochs with batch size of $64$ on a NVIDIA GeForce RTX 2080Ti GPU. We set event channels $B$ to $3$, number of experts $N_e$ to $8$ and $TopK$ to $2$. 
 We set the number of input frames/tensors $m$ to $2$. 
We implemented  MCO-E Net in PyTorch. We trained AdamW with a weight decay $0.001$, and the learning rate was set to $0.0003$.

\subsection{Comparison with State-of-the-art Methods}
\begin{figure}
    \centering
    \includegraphics[width=\linewidth]{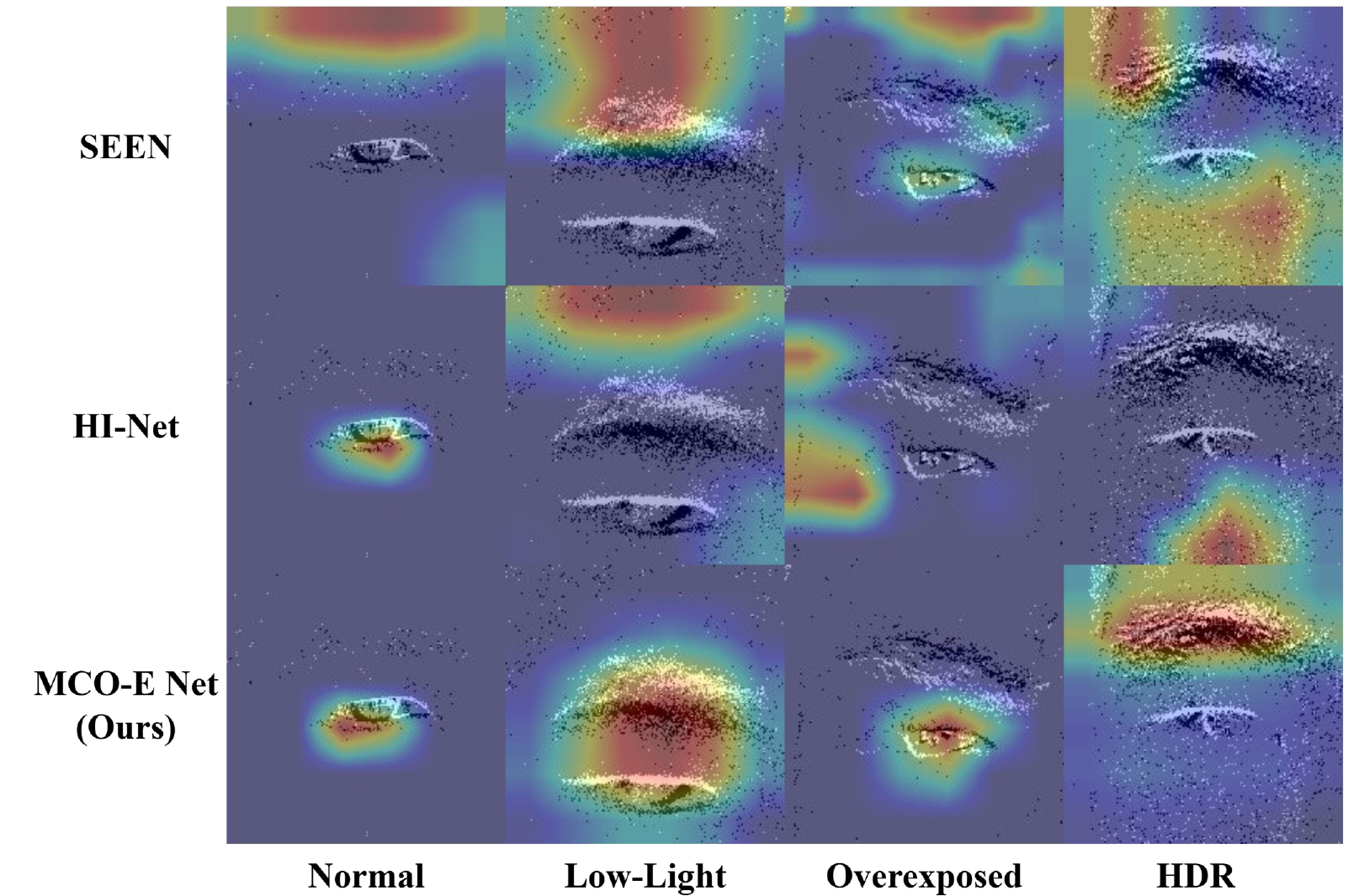}
    \caption{Heatmap visualization of the comparison between our proposed MCO-E Net and SOTA methods}
    \label{fig:SOTA_vis}
\end{figure}
Since combining RGB and Event in Single-eye expression recognition task is a new strategy, related work is scarce. 
Therefore, to prove the performance of our network, our network is compared with RGB-based methods: 
EMO~\cite{wu2020emo}, Eyemotion~\cite{Hickson2017EyemotionCF}, 
Former DFER~\cite{zhao2021former}, R(2+1)D~\cite{tran2018closer}, 
3D Resnet18~\cite{hara2018can}, Resnet50+GRU~\cite{cho2014learning}, Resnet18+
LSTM~\cite{6795963}
and Event-based methods: 
SEEN~\cite{Zhang_2023}, 
MSKD~\cite{ijcai2024p350} and
HI-Net~\cite{Han2024HierarchicalEI}. 
In addition, the expression recognition method can be divided into eye judgment and entire face judgment, as shown in Table.~\ref{tab:SOTA},  the comparison methods are divided into Face and Eye. 
To ensure a rigorous and impartial assessment of model performance while enabling equitable comparisons with SOTA methods, we employ two standardized evaluation metrics: UAR  and WAR. And Accuracy under four lighting conditions: Normal, Overexposure, Low-Light and HDR, Accuracy in seven different emotions: Happy, Sadness, Anger, Disgust, Surprise, Fear and Neutral. 

As shown in Table.~\ref{tab:SOTA}, on  SEE dataset, 
the performance of our proposed method is significantly higher than that of existing SOTA  methods in WAR and UAR. 
Specifically, our method achieved $91.3\%$ and $91.9\%$ in WAR and UAR respectively. In WAR and UAR metrics, surpass the current state-of-the-art methods $4.4\%$ and $4.2\%$. 
Besides, we also show performance under four different lighting conditions.  As demonstrated in Table.~\ref{tab:SOTA}, our method shows the best accuracy in all lighting conditions. 
We achieve best accuracy in Normal, Overexposure, Low-Light and HDR conditions of $87.7\%$, $93.8\%$, $93.6\%$ and $90.2\%$ respectively, and it outperforms the runner-up by $3.1\%$, $3.5\%$, $5.3\%$and $5.0\%$.  
In addition, as shown in Fig.~\ref{fig:SOTA_vis}, our method demonstrates superior capability in capturing and emphasizing semantic features around eye or eyebrow regions across four lighting conditions when compared with SEEN~\cite{Zhang_2023} and HI-Net~\cite{Han2024HierarchicalEI}. 
This focused perception enhances expression classification accuracy.

\subsection{Ablation study}
\begin{figure}
    \centering
    \includegraphics[width=\linewidth]{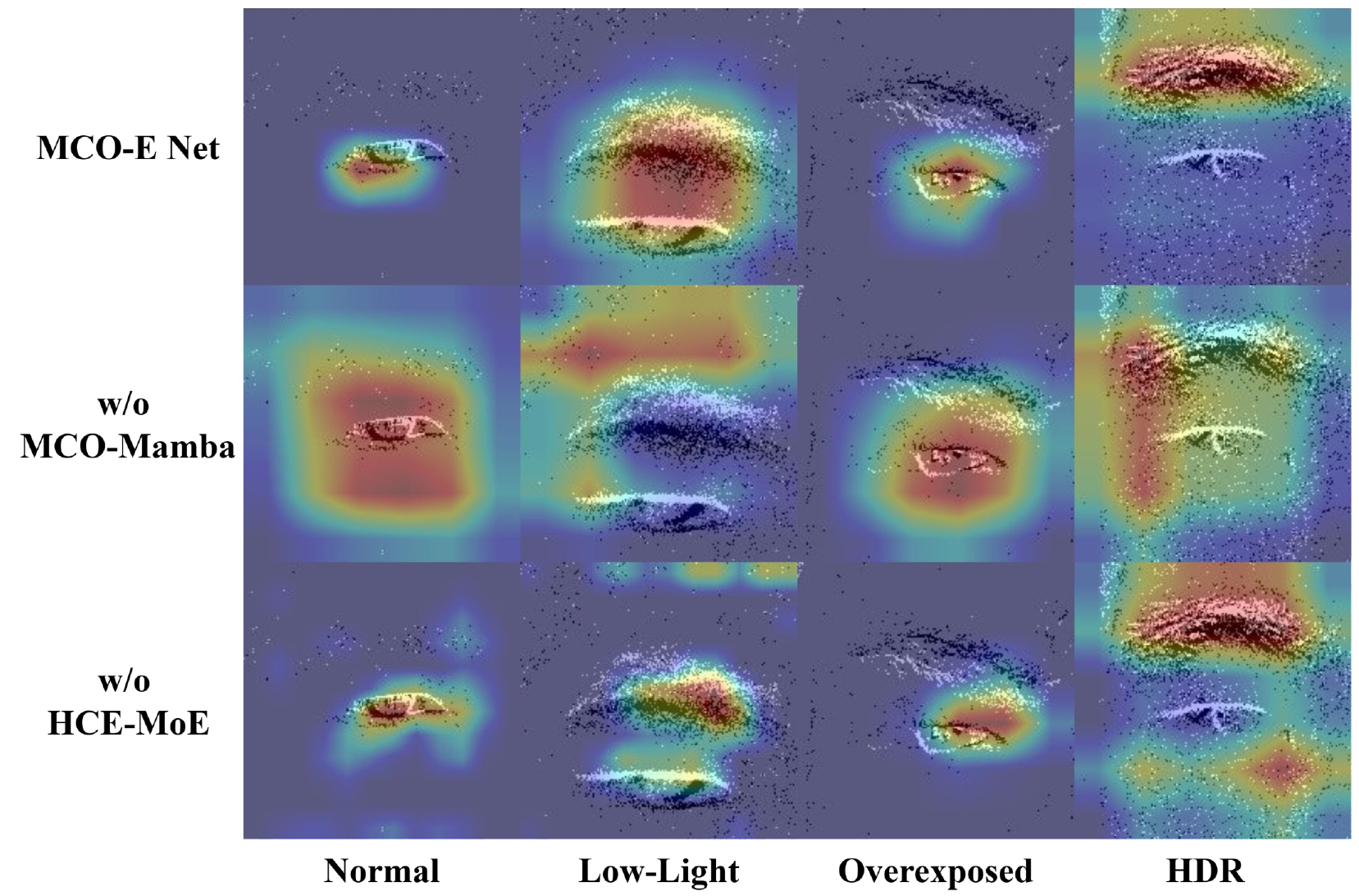}
    \caption{Heatmap visualization of the comparison between our proposed MCO-E Net and removing MCO-Mamba or HCE-MoE}
    \label{fig:abaltion_vis}
\end{figure}

\begin{table}[h!]
\centering
\caption{Results produced by combining different components of our proposed network. }
\label{tab_ablation1}
\begin{tabular}{l|cc}
\toprule[1.2pt]
Methods    & WAR & UAR \\ \hline
    \textit{A. }RGB Only & 83.5 & 84.3  \\
    \textit{B. }Event Only & 72.6 & 73.3 \\ \hline        
    \textit{C. }w/o MCO-Mamba & 89.4 & 90.1\\ 
    \textit{D. }MCO-Mamba(w/o MJOS)& 89.9 & 90.5 \\
    \textit{E. }MCO-Mamba(w/o MCIB)& 89.8 & 90.5\\ \hline
    \textit{F. }w/o HCE-MoE & 88.9 & 89.7\\    
    \textit{G. }HCE-MoE (1 type of expert) & 89.9 & 90.5 \\  
    \textit{H. }HCE-MoE (2 type of experts) & 90.6 & 91.3\\ \hline
    \textit{I. }Ours& 91.3 & 91.9 \\                 \bottomrule[1.2pt]
\end{tabular}
\end{table}

\subsubsection{Effectiveness of Multi-modality Semantics. }
The experimental results are displayed on Table.~\ref{tab_ablation1}, referred to as $A$ and $B$, to verify the effectiveness of multi-modality. 
Experiment $A$: RGB Only, that is, only RGB data is used as input. 
Experiment $B$: Event Only, that is, only event data is used as input. 
From experiments $A$ and $B$ in Table.~\ref{tab_ablation1}, it can be clearly seen that the method of using multi-modality as input is significantly better than the method of rely solely on RGB or Event. 
The event modality preserves high-temporal-resolution motion patterns around the eyes, demonstrating robustness against extreme lighting condition, while the RGB modality maintains rich spatial-semantic features around  eyes.

\subsubsection{Effectiveness of MCO-Mamba }
In proposed MCO-Mamba, we use MJOS (for Mamba) and MCIB block for joint optimization of model and Multi-modal collaborative modeling, respectively.  
To prove the efficacy of MCO-Mamba, we design three ablation experiment \textit{C} to \textit{E} of Table.~\ref{tab_ablation1}: 
(i) w/o MCO-Mamba, remove MCO-Mamba from the network. 
(ii) MCO-Mamba(w/o MJOS), remove MJOS  from MCO-Mamba, only the MCIB block is retained. 
(iii) MCO-Mamba(w/o MCIB), remove MCIB from MCO-Mamba, directly add the features processed by MJOS.  

From the experimental results shown in \textit{C} to \textit{E} of Table.~\ref{tab_ablation1}, we can analyze the most impact on performance is row \textit{C}. 
To the  removal of MCO-Mamba,  the advantageous semantics of these two modalities cannot be combined, resulting in performance degradation. 
\textit{D} of Table.~\ref{tab_ablation1} is remove  MJOS  from MCO-Mamba. 
This means that  the features extracted by backbone are directly fusion through  MCIB block. 
The lack of parameter joint  optimization  strategy leads to significant reduction in model performance. 
\textit{E} of Table.~\ref{tab_ablation1} is remove MCIB from MCO-Mamba, change the fusion method to feature addition.
The lack of semantic integration leads to performance degradation in WAR and UAR.

\subsubsection{Effectiveness of HCE-MoE. }
For HCE-MoE, we performed three experiment: 
(i) w/o HCE-MoE, remove the HCE-MoE in the network and change to the full connection layer. 
(ii) HCE-MoE(1 type of expert), only the deep experts are retained in HCE-MoE. 
(iii) HCE-MoE(2 type of experts), retain deep experts and attention experts in HCE-MoE. 
By analyzing the \textit{F} to \textit{H} of Table.~\ref{tab_ablation1}, we can be seen that the importance of HCE-MoE to performance. 
First, as shown in \textit{F} of Table.~\ref{tab_ablation1}, we replace the HCE-MoE with a fully connected layer, and the performance was significantly reduced, with WAR and UAR dropping by $2.4\%$ and $2.2\%$ respectively. 
This is because, removing HCE-MoE, the model loses the ability of multiple experts to process features, different experts can focus on different sub-regions or feature patterns of input data (for example, some experts deal with edge features, others deal with texture features), and achieve task decoupling through dynamic routing of gated networks to improve the fine-grained classification. 
Second, as shown in \textit{G} and \textit{H} of Table.~\ref{tab_ablation1}, the performance of both retaining one type of expert and retaining two type of experts has decreased compared to our HCE-MoE. This means that having more type experts leads to higher pattern diversity and reduces the likelihood of key feature patterns being overlooked. 
In addition, From Fig.~\ref{fig:abaltion_vis} we can also see that when MCO-Mamba or HCE-MoE is removed, the semantics perceived by the model is not as focused as our method, which is not conducive to emotion classification under extreme lighting conditions.

\subsection{Discussion of MCO-Mamba}
\begin{table}[h]
\centering
\caption{Results produced by different SSM matrix Interaction of our proposed MCO-Mamba. }
\label{tab:ablation2}
\begin{tabular}{l|cc}
\toprule[1.2pt]
Methods    & WAR & UAR \\ \hline
\textit{A. }MCO-Mamba ($A$)   &  90.2 & 90.8      \\
\textit{B. }MCO-Mamba ($B$) &   90.2 & 90.8  \\ 
\textit{C. }MCO-Mamba ($C$)   &  90.1 & 90.7   \\ 
\textit{D. }MCO-Mamba ($D$) &    89.8 & 90.5     \\ \hline

\textit{E. }MCO-Mamba ($A,B$)   & 89.9 & 90.6   \\ 
\textit{F. }MCO-Mamba ($A,C$)   & 90.2 & 90.9   \\ 
\textit{G. }MCO-Mamba ($A,D$)   & 89.5 & 90.2   \\ 
\textit{H. }MCO-Mamba ($B,D$)   & 89.8 & 90.4   \\
\textit{I. }MCO-Mamba ($C,D$)   & 90.1 & 90.7   \\ \hline
\textit{J. }MCO-Mamba ($A,B,C,D$) & 90.0 & 90.6   \\ \hline
\textit{K. }MCO-Mamba ($B,C$)       &91.3 &91.9 \\ \bottomrule[1.2pt]
\end{tabular}
\end{table}

To verify the effectiveness of MCO-Mamba, we designed experiments from three perspectives: parameter sharing, selection of interactive parameters, and exchange of parameters.

\textbf{Parameter sharing. }
As mentioned in Sub.~\ref{sec:related}, static interaction strategies (such as simple parameter sharing or exchanging)
that force a unified feature space tend to weaken modality-specific
features and cause an imbalance between modality-shared and
modality-specific features. 
To demonstrate that the static strategy that only sharing parameters between modalities affects modality-specific representations, we designed experiments $\textit{A}$ to $\textit{D}$ of Table.~\ref{tab:ablation2}. 
$\textit{A}$ to $\textit{D}$ respectively indicate that the Multi-modal Joint Optimization Scheme (MJOS) is removed in our MCO-Mamba, and only a single parameter matrix $A$, $B$, $C$ or $D$ is shared. 
From the results, we can see that the performance drops dramatically. 
This is because the shared SSM parameter matrix affects the modality-specific representation, resulting in an imbalance in the representation between modalities.

\textbf{Selection of interactive parameters. }
As mentioned in Subsec.~\ref{subsub:PIMamba}, to validate our choice of interacting $B$ and $C$, we designed experiments $E$ to $I$ of Table.~\ref{tab:ablation2}, which represent interactive operations $\mathcal{M}$ on different parameter matrices. 
From the analysis $E$ to $I$ of Table.~\ref{tab:ablation2}, we can see that the performance has dropped significantly due to the $A$ and $D$. 
$A$ interacts between two modalities, resulting in unstable state transitions. 
$D$ is a residual term, and interacting with $D$ will affect the introduction of original information. 

\textbf{Exchange of parameters. } As shown in row $J$ of Table.~\ref{tab:ablation2}, we exchange all SSM parameters of the two modalities. 
The observed performance degradation stems from the oversimplified exchange mechanism undermining the critical modal-sharing features, which are essential for maintaining synergistic coupling between multi-modality.

\subsection{Discussion of HCE-MoE}
\begin{table}[h]
\centering
\caption{Results produced by change the router}
\label{tab:ablation5}
\begin{tabular}{l|cc}
\toprule[1.2pt]
Methods    & WAR & UAR \\ \hline
\textit{A. }MLP   &    88.9 & 89.7    \\
\textit{B. }Deep experts Only.   &    90.2 & 90.9    \\
\textit{C. }Attention Experts Only   &    89.9 & 90.6    \\ 
\textit{D. }Focal Experts Only   & 89.9 & 90.5   \\ 
\hline
\textit{E. }Ours   &91.3 &91.9 \\ \bottomrule[1.2pt]
\end{tabular}
\end{table}
The ablation study on router architectures as shown in Table.~\ref{tab:ablation5}, where we designed four experiment $A$ to $D$: 
$A$ is to exchange HCE-MoE for MLP;
$B$ is to keep only Deep expert in HCE-MoE; 
$C$ is to keep only Attention Experts in HCE-MoE; 
$B$ is to keep only Focal Experts in HCE-MoE. 
our proposed routing mechanism $E$ achieves state-of-the-art performance. 
The MLP baseline ($\textit{A}$) produces the weakest results, and our designed expert show clear advantages: the Deep Expert ($\textit{B}$) achieves 90.2\% WAR, slightly better than the Attention Expert ($\textit{C}$) and Focal Expert ($\textit{D}$), indicating that the deep expert provides slightly greater discriminative power for this task.

\begin{table}[h]
\centering
\caption{Results produced by change the number of experts of HCE-MoE}
\label{tab:ablation3}
\begin{tabular}{l|cc}
\toprule[1.2pt]
Methods    & WAR & UAR \\ \hline
\textit{A. }$N_e=5$   &  90.0 & 90.6   \\
\textit{B. }$N_e=10$   &   89.0 & 89.7   \\ 
\textit{C. }$N_e=15$   &   89.5 & 90.3    \\ \hline
\textit{D. }Ours ($N_e=8$)       &91.3 &91.9 \\ \bottomrule[1.2pt]
\end{tabular}
\end{table}
Table.~\ref{tab:ablation3} demonstrates the impact of varying the number of experts ($N_e$) in our HCE-MoE architecture on WAR and UAR. 
While configuration $A$ ($N_e=5$) achieves metrics of 90.0\% WAR and 90.6\% UAR, increasing the expert count to $10$ (configuration $B$) paradoxically degrades performance to 89.0\% WAR and 89.7\% UAR. Subsequent expansion to $15$ experts (configuration $C$) yields partial recovery (89.5\% WAR, 90.3\% UAR), yet still underperforms relative to the baseline. 
The result demonstrating that intermediate expert counts enable more effective knowledge specialization. This optimal balance suggests that: Insufficient experts limit model capacity for capturing complex pattern variations; Excessive experts introduce parameter complexities, resulting in reduced performance.

\begin{table}[h]
\centering
\caption{Results produced by change the $Topk$ of HCE-MoE}
\label{tab:ablation4}
\begin{tabular}{l|cc}
\toprule[1.2pt]
Methods    & WAR & UAR \\ \hline
\textit{A. }$k=1$   &   89.6 & 90.3   \\
\textit{B. }$k=3$   &   90.3 & 90.9   \\ 
\textit{C. }$k=5$   &   90.4 & 91.0   \\ \hline
\textit{D. }Ours ($k=2$)       &91.3 &91.9 \\ \bottomrule[1.2pt]
\end{tabular}
\end{table}
The experimental results in Table.~\ref{tab:ablation4} demonstrate a key balance in expert activation for our HCE-MoE framework. 
While extending the activation experts from $k=1$ to $k=3$ improves recognition accuracy, further upgrading to $k=5$ yields only marginal gains, suggesting the inherent limitations of indiscriminate expert aggregation. 
Our configuration with $k=2$ achieves optimal performance, outperforming both under activated and over activated settings by significant margins.

\section{CONCLUSION}
In this paper, we proposed a Multi-modal Collaborative Optimization and Expansion Network (MCO-E Net), for the single-eye expression recognition tasks. The MCO-E Net contains two novel designs:  Multi-modal Collaborative Optimization  Mamba (MCO-Mamba),  Heterogeneous Collaborative and Expansion MoE (HCE-MoE). 
The MCO-Mamba  drived the model to balance the learning of two-modal semantics and capture the advantages of both modalities through joint optimization in Mamba. 
The HCE-MoE systematically combines multiple feature expertise through a heterogeneous architecture, enabling collaborative learning of complementary semantics and capturing comprehensive semantics. 
Extensive experiments demonstrate that our MCO-E Net achieves competitive  performance on the single-eye expression recognition task. 


\clearpage
\bibliographystyle{ACM-Reference-Format}
\bibliography{sample-base}


\end{document}